\def\year{2021}\relax
\documentclass[letterpaper]{article} 
\usepackage{aaai21}  
\usepackage{times}  
\usepackage{helvet} 
\usepackage{courier}  
\usepackage[hyphens]{url}  
\usepackage{graphicx} 
\usepackage{natbib}  
\usepackage{caption} 
\usepackage{multirow}
\usepackage{amssymb} 
\usepackage{booktabs}
\usepackage{bm}
\usepackage{CJKutf8}
\usepackage[switch]{lineno}
\newcommand{\tabincell}[2]{\begin{tabular}{@{}#1@{}}#2\end{tabular}}

\title{Label Enhanced Event Detection with Heterogeneous Graph Attention Networks}

\author{Shiyao Cui$^{1,2}$  \ Bowen Yu$^{1,2}$ Xin Cong$^{1,2}$ \ Tingwen Liu$^{*,1,2}$ \  Quangang Li$^{1,2}$ \ Jinqiao Shi$^{1,3}$ \\}
\affiliations{
    $^1$Institute of Information Engineering, Chinese Academy of Sciences. Beijing, China
    \\
    $^2$School of Cyber Security, University of Chinese Academy of Sciences. Beijing, China \\
    $^3$Beijing University of Posts and Telecommunications. Beijing, China 
    \\
    \{cuishiyao, yubowen, congxin, liutingwen, liquangang\}@iie.ac.cn
    \\
    shijinqiao@bupt.edu.cn
}

\begin{document}
\maketitle

\begin{abstract}
Event Detection (ED) aims to recognize instances of specified types of event triggers in text.
Different from English ED, Chinese ED suffers from the problem of word-trigger mismatch due to the uncertain word boundaries. 
Existing approaches injecting word information into character-level models have achieved promising progress  to alleviate this problem, but they are limited by two issues.
First, the interaction between characters and lexicon words is not fully exploited.
Second, they ignore the semantic information provided by event labels. 
We thus propose a novel architecture named Label enhanced Heterogeneous Graph Attention Networks (L-HGAT). 
Specifically, we transform each sentence into a graph, where character nodes and word nodes are connected with different types of edges, so that the interaction between words and characters is fully reserved.
A heterogeneous graph attention networks is then introduced to propagate relational message and enrich information interaction.
Furthermore, we convert each label into a trigger-prototype-based embedding, and design a margin loss to guide the model distinguish confusing event labels.
Experiments on two benchmark datasets show that our model achieves significant improvement over a range of competitive baseline methods.
\end{abstract}

\section{Introduction}
Event Detection (ED),  the task of which involves identifying the boundaries of event triggers and classifying them into the corresponding event types, aims to seek recognize events of specific types from given texts.
As a fundamental task of information extraction, many high-level NLP tasks, such as information retrieval~\cite{dart2014wsBasile} and question answering~\cite{DBLP:conf/sigir/YangCWK03}, need an event detector as one of their essential components.
\begin{figure}[!t]
	\centering
	\includegraphics[width=0.85\columnwidth]{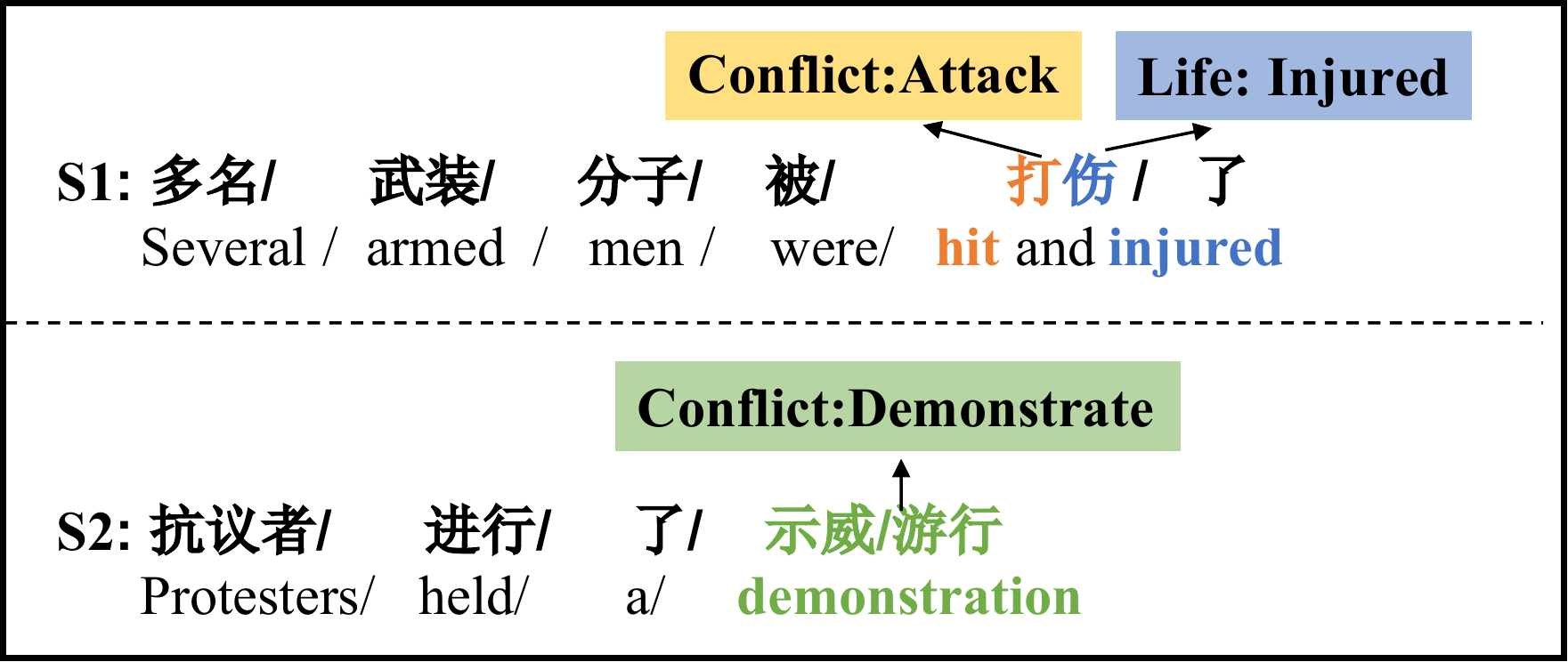}
	\caption{An example of word-trigger mismatch problem.}
	\label{fig:example}
\end{figure}
\begin{figure}[!h]
	\centering
	\includegraphics[width=0.85\columnwidth]{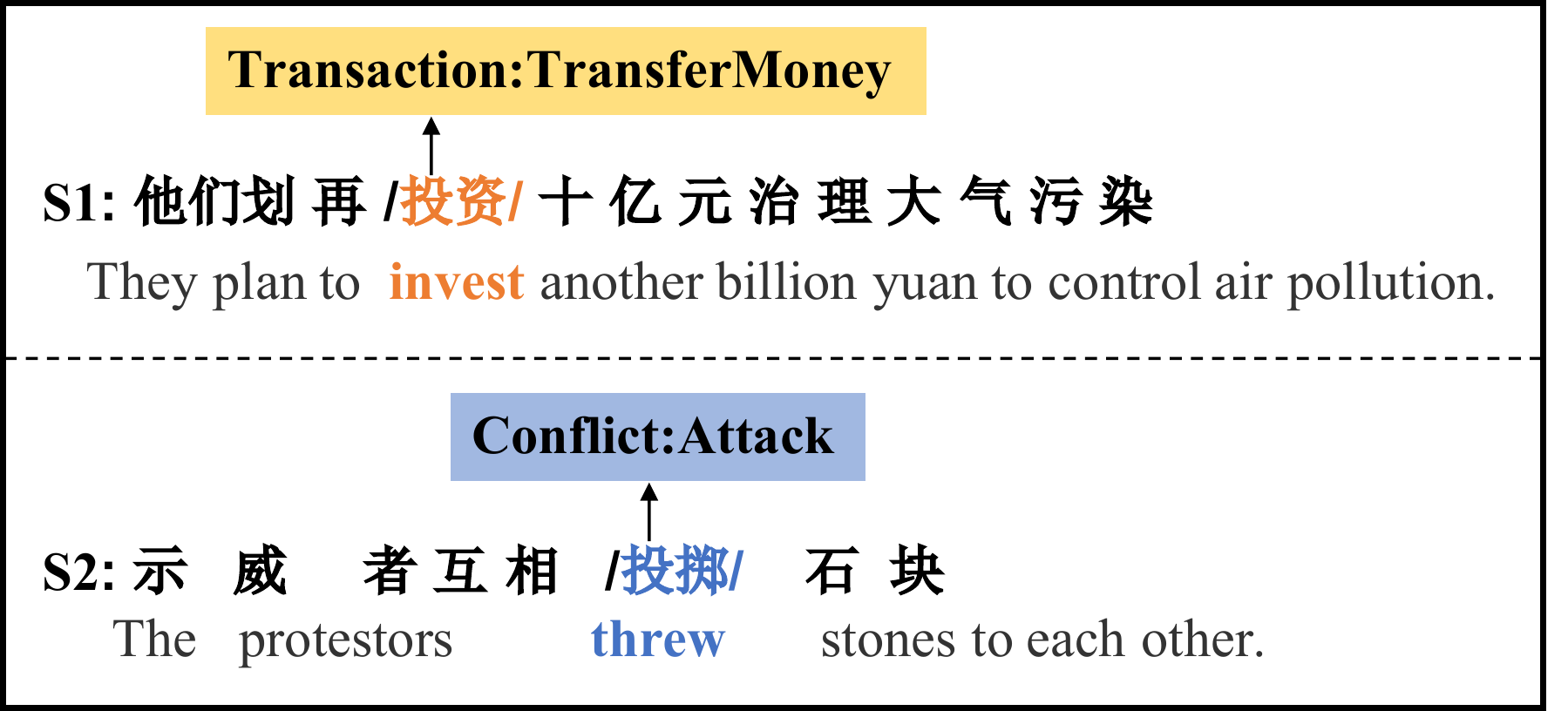}
	\caption{An example of ambiguous semantics.}
	\label{fig:example2}
\end{figure}

Recent studies~\cite{chen-etal-2015-event,nguyen-etal-2016-joint-event,liu-etal-2017-exploiting,liu-etal-2018-jointly,yan-etal-2019-event,cui-etal-2020-edge} show that English ED models have achieved great performance by treating the problem as a word-by-word sequence labeling task. 
Different from English ED, many East Asian languages, including Chinese, are written without explicit word boundary, resulting a much tricky ED task.
An intuitive solution is to apply Chinese Word Segmentation (CWS) tools first to get word boundaries, and then use a word-level sequence labeling model similar to the English ED models. 
However, word boundary is ambiguous in Chinese thus word-trigger mismatch problem exists in Chinese ED, where an event trigger may not exactly match with a word, but is likely to be part of a word or cross multiple words as Figure~\ref{fig:example} demonstrates.
Meanwhile, character-level sequence tagging is able to alleviate this problem, but Chinese character embedding can only carry limited information due to the lack of word and word-sequence information, resulting to ambiguous semantics.
\begin{CJK}{UTF8}{gbsn}
	For example in Figure~\ref{fig:example2}, the character ``投''  in lexicon word ``投资(invest)'' and ``投掷(throw)'' has entirely different meanings, triggering the event of ``Transaction:TransferMoney'' and ``Conflict:Attack'', respectively.
\end{CJK}

Several recent works have demonstrated that considering the lexicon word information could provide more exact information to discriminate semantics of characters.
\citeauthor{lin-etal-2018-nugget}~\shortcite{lin-etal-2018-nugget} designed NPN, a CNN-like network to model character compositional structure of trigger words and introduced a gate mechanism to fuse information from characters and words.
~\citeauthor{ding-etal-2019-event}~\shortcite{ding-etal-2019-event} proposed TLNN, a trigger-aware Lattice LSTM architecture, exploiting semantics from matched lexicon words to improve Chinese ED.

Although these methods~\cite{lin-etal-2018-nugget, ding-etal-2019-event} have achieved great success, they continue to have difficulty in fully exploiting the interaction between characters and lexicon words.
Specifically, for each character, NPN exploits a gate mechanism to fuse its information with one corresponding word.
This means that each character could only be incorporated with one matched word, but actually one character is likely to match with several words, leading to information loss.
For TLNN, it constructs cut paths to link the start and end character for each matched word, but semantic information of the matched lexicon word fails to flow into all the characters it covers except the last one, due to the inherently unidirectional sequential nature of Lattice LSTM~\cite{sui-etal-2019-leverage, Mengge2019PorousLT}.

Besides, previous ED works usually ignore semantic information maintained by the event types.
We observe that event types are usually semantically related to the corresponding event triggers. 
\begin{CJK}{UTF8}{gbsn}
	For example, some common event triggers of type ``Conflict:Attack(攻击)'', such as ``hit(击打)'', ``strike(撞)'' and ``invade(侵略)'', are specific behaviors of ``Attack''.
\end{CJK}
Such an observation shows that considering the semantic information of event labels may provide fine-grained semantic signals to guide the detection of event triggers, and accordingly benefit ED performance.

In this paper, we propose a novel neural architecture, named Label Enhanced Heterogeneous Graph Attention Networks (L-HGAT), for Chinese ED.
To promote better information interaction between words and characters, we transform each sentence into a graph. 
We first connect lexicon words with all the characters it covers.
And then neighboring characters are also linked with each other to provide local context information to enhance character representations, especially for those without matched lexicon word.
To capture different granularity of semantic information from words and characters, we formulate words and characters as two types of nodes, thus a heterogeneous graph attention networks is utilized to enable rich information propagation over the graph.
Additionally, we design a matcher module to leverage the semantic information of event labels.
Specifically, we transform event labels into an event-trigger-prototype based embedding matrix by summarizing the trigger representations belonging to each event label.
Based on the generated event label representation, a margin loss is further exploited to enhance the ability to discriminate confusing event labels.
Comparing with previous works, our contributions are as follows:
\begin{itemize}
	\item To the best of our knowledge, we are the first to utilize heterogeneous graph attention networks, incorporating different types of nodes, to enhance message passing between characters and lexicon words for Chinese ED task.
	\item As far as we know, we are the first to mine the interaction between event triggers and labels to guide the recognition of event triggers in standard Chinese ED paradigm.
	\item Our model consistently achieves superior performance over previous competing approaches on two benchmarks, ACE2005 and KBP2017. Further analysis confirms the effectiveness of our model.
\end{itemize}

\section{Related works}
\subsection{Chinese Event Detection}
Different from English ED, Chinese event triggers are more difficult to be recognized due to the ``word-trigger mismatch'' problem.
Transforming ED into a character-wise sequence labeling paradigm helps to alleviate the problem, but characters contain only limited semantic information, and lexicon words could provide more exact information to discriminate semantics of characters.
Consequently, character-wise models incorporated with word information have attracted research attention.
\citeauthor{lin-etal-2018-nugget}~\shortcite{lin-etal-2018-nugget} proposed NPN neural network to learn a hybrid representation fused with information from characters and words.
Recently, \citeauthor{ding-etal-2019-event}~\shortcite{ding-etal-2019-event} designed TLNN, a trigger-aware Lattice LSTM structure, to incorporate lexicon word to improve ED.

These works have achieved great progress, but they still face two issues.
First, they do not fully explore interaction between characters and words.
For NPN, each character could only be incorporated with one matched word, which may lose richer information from other matched words.
For TLNN, previous works~\cite{sui-etal-2019-leverage, Mengge2019PorousLT} have pointed out that the undirectional sequential structure of Lattice LSTM limits the information flow from lexicon word to all the characters it covers except the last one.
Besides, we notice that previous event detection works ignore the semantic information maintained by event labels,  which may lose potential classification indicators to improve ED.

\subsection{Heterogeneous Graph for NLP}

Graph Neural Networks~\cite{kipf2017semi} is originally designed for homogeneous graph, where all nodes share the same type.
However, graphs in real scenarios are usually equipped with multiple types of nodes and edges, thus heterogeneous graph neural network (HGCN)~\cite{Zhang2019HeterogeneousGN} and HGAT (an improved version of HGCN with attention mechanism)~\cite{han2019} are proposed.
Recent works have exploited HGCN or HGAT in many NLP fields, such as multi-hop reading comprehension~\cite{tu-etal-2019-multi}, semi-supervised short text classification~\cite{linmei-etal-2019-heterogeneous} and summarization~\cite{wang-etal-2020-heterogeneous}.

For event detection task, GCN-based methods~\cite{nguyen2018graph, liu-etal-2018-jointly, yan-etal-2019-event, cui-etal-2020-edge} have been successfully deployed for English ED.
They consider words as nodes and construct a homogeneous graph, which is insufficient for Chinese ED to fully exploit characters and words.
Additionally, these works construct graph based on syntactic dependency  parsed from external tools like Stanford CoreNLP toolkit, which may suffer from error propagation.

\noindent In this paper, we propose a HGAT-based model for Chinese ED by formulating characters and matched lexicon words as two different types of nodes, and construct graph without relying on external tools.
Besides, we mine the semantic clues provided by event labels to guide the recognition of event triggers.
Although some works~\cite{huang-etal-2018-zero, lai-nguyen-2019-extending, Du2020EventEB} have tried to make use of semantics of event labels, it should be noted our method differs with theirs in two aspects.
First, we use event labels in different scenarios.
\citeauthor{huang-etal-2018-zero}\shortcite{huang-etal-2018-zero} works on Zero-Shot event extraction task, \citeauthor{lai-nguyen-2019-extending}\shortcite{lai-nguyen-2019-extending} focuses on discovering new event types, while our work mines the interaction between triggers and event labels in standard Chinese ED paradigm. 
Second, we use and obtain semantic information of event labels in different ways. \citeauthor{huang-etal-2018-zero}\shortcite{huang-etal-2018-zero} learns event type representation with event ontology consisting of its roles, \citeauthor{lai-nguyen-2019-extending}\shortcite{lai-nguyen-2019-extending} defines event type as a set of keywords to find new event types, \citeauthor{Du2020EventEB}\shortcite{Du2020EventEB}  designs question template based on the semantics of triggers, while we use trigger-prototype-based embedding as event label embedding and fine-tune it. 

\section{Problem Statement}
We formulate event detection as a character-wise sequence labeling task, where each character is assigned a label to decide whether it is in relevant to an event trigger.
Label ``O'' means the character is independent of target event label.
Other labels are formatted as ``B-EventType" and ``I-EventType", which respectively means that the character is the beginning character, inside character of an event trigger.
Therefore, the total number of event labels is $2 \times N_{e} + 1$, where $N_{e}$ is the number of predefined event types.

\section{Method}
\begin{figure*}
	\centering
	\includegraphics[width=0.95\linewidth]{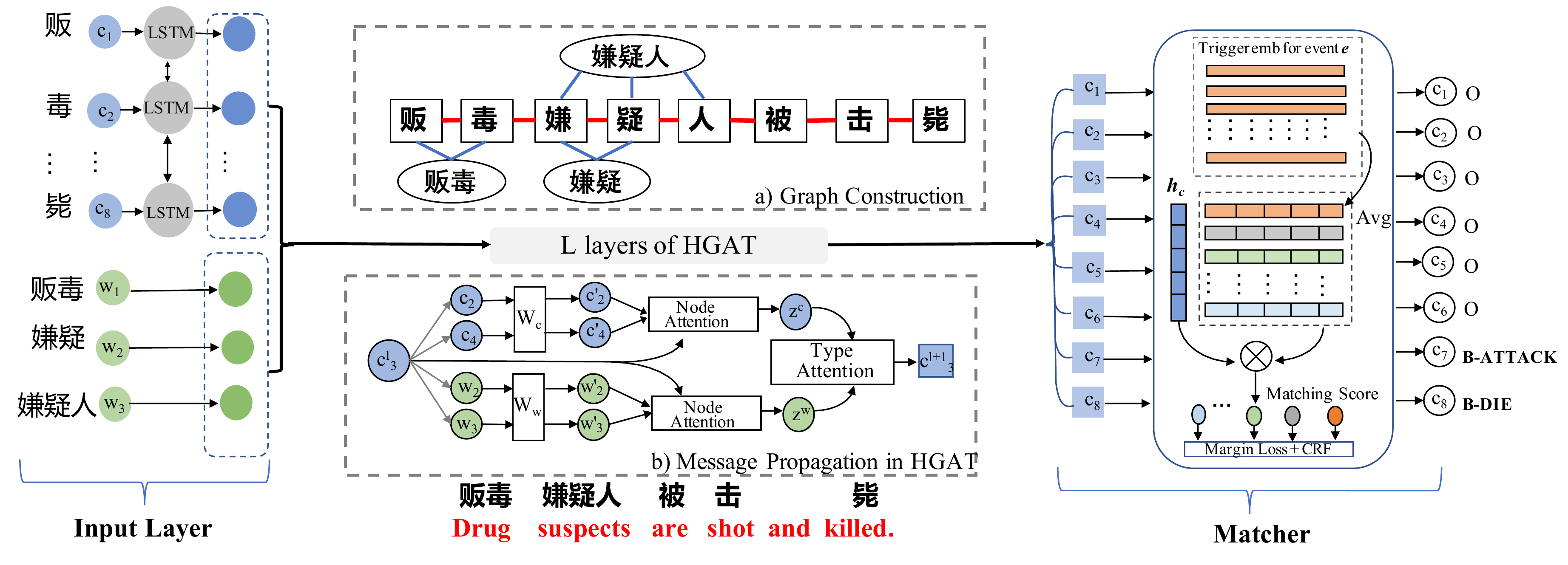}
	\caption{Illustration of our proposed architecture, which consists of an Input Layer, $L$ layers of HGAT and a matcher module. Subfigure a) and b) show the details of graph construction and message propagation process in HGAT.}
	\label{fig:model}
\end{figure*}
In this section, we first introduce the construction and initialization of our heterogeneous graph, based on which the heterogeneous graph attention networks is employed to fully integrate information between words and characters.
Then we detail the design of our matcher module, which leverages the event label semantics to guide the recognition of triggers, and employs CRF loss and margin loss for model training.

\subsection{Graph Construction}

For an $n$-character Chinese sentence $S=\{c_1, c_2, ..., c_n\}$, it could be denoted as a word sequence as $S_w = \{w_{(b_1,e_1)}, w_{(b_2,e_2)}, ..., w_{(b_m,e_m)}\}$, where $b_i$ and $e_i$ are respectively the index of beginning and ending character which the $i_{th}$ word matches in $S$.
\begin{CJK}{UTF8}{gbsn}
	For example, in Figure~\ref{fig:model}, $w_1$=``贩毒(selling drugs)'' with $b_1=1$ pointing to ``贩(selling)''  and $e_1=2$ pointing ``毒(drugs)''.
\end{CJK}
Each sentence is transformed into a heterogeneous graph, with two types of nodes (characters, words) and three kinds of edges.
The first kind of edge named ``c2c-edge'' connects neighboring characters, which incorporates local context information for characters to enrich semantic information.
The second kind of edge named ``w2c-edge'' connects lexicon words with containing characters, enabling information from words flow into all characters it covers.
The third kind of edge named ``c2w-edge'' is the reverse of ``w2c-edge'', allowing information propagation from characters to words, thus sequential context semantics could be injected into words.
Due to the heterogeneity of nodes, semantic information from different granularity levels could be learned and fused.


\subsection{Input Layer(Graph Initializer)}
We first transform each character and matched lexicon words into real-valued embeddings by individually looking up a pre-trained character embedding and word embedding matrix.
Let $ \mathbf{X}_{c}\in \mathbb{R}^{n \times d}$ and $ \mathbf{H}_{w} \in \mathbb{R}^{m \times d}$ represent the input embedding matrix of character sequence $S$ and word sequence $S_w$, where $d$ is the embedding dimension,  $n$ and  $m$ are respectively the number of characters and matched words in the sentence.
A BiLSTM layer is then adopted to capture the sequential context information for each character in the sentence.
By concatenating the forward and backward LSTM hidden states, we obtain the contextual representations as $\mathbf{H}_c = \{h_1, h_2, ..., h_n\}$, where $\mathbf{H}_c \in \mathbb{R}^{n \times d}$.
Moreover, $\mathbf{H}_c$ and $\mathbf{H}_w$ are used as initial node features in HGAT.

\subsection{Heterogeneous Graph Attention Neural Networks}

Given a graph with two kinds of nodes and three kinds of edges, we leverage HGAT to enable information propagation along the graph.
Heterogeneous graph convolution is exploited to aggregate information from different types of nodes.
Furthermore, considering that different neighboring nodes and node types have different effects on a specific node, the attention mechanism is explored to aggregate the information from different types of neighbors.

\subsubsection{Node Attention}

GAT~\cite{velickovic2018graph} is an improved version of GCN, which exploits node-level attention to reduce the weight of noisy neighboring nodes.
Unlike vanilla GAT, node-type specific transformation matrix is used to project different types of node feature into the same feature space considering the heterogeneity of nodes.
For a $\tau$-type node $\mathbf{h}_{j}^l$ in the $l_{th}$ layer of HGAT, the projection process is shown as follows:
\begin{equation}
\mathbf{\hat{h}}_{j}^{l} = \mathbf{W}_{\tau} \mathbf{h}_{j}^l.
\end{equation}
The attention mechanism is then exploited to learn the association between node pairs and conduct node aggregation over the graph as follows:
\begin{equation}
\mathbf{e}_{ij} = {\rm{LeakyReLU}} (\mathbf{v}^{\tau} [\mathbf{\hat{h}}_{i}^{l}, \mathbf{\hat{h}}_{j}^{l}]) ,
\end{equation}
\begin{equation}
a_{ij} = \frac{ {\rm{exp}} (\mathbf{e}_{ij})}{\sum\limits_{j \in N_{\tau,i}}  {\rm{exp}}(\mathbf{e}_{ij})}  ,
\end{equation}
\begin{equation}
\mathbf{z}_{i}^{\tau} = \sigma(\sum\limits_{j \in N_{\tau,i}}^{n}a_{ij} \mathbf{\hat{h}}_{j}^{l}) ,
\end{equation}
where $\mathbf{W}_{\tau}$, $\mathbf{v}_{\tau}$ are trainable weights, $N_{\tau,i}$ is the set of $\tau$-type neighboring nodes of $\mathbf{h}_{i}^{l}$, $\mathbf{z}_{i}^{\tau}$ is the semantic embeddings from $\tau$-type neighboring nodes of $\mathbf{h}_i^l$.

\subsubsection{Type Attention}

For a character node $\mathbf{h}_i^l$, we can obtain two types of semantic embedding, $\mathbf{z}_{i}^{c}$ and  $\mathbf{z}_{i}^{w}$, by respectively operating GAT over neighboring character nodes and word nodes.
In order to fuse semantic embeddings from character granuality and word granuality, a type-level attention mechanism is designed to generate a comprehensive representation $\mathbf{h}_i^{l+1}$ for the next layer.
Specifically, we weight semantic embedding from different types of neighbors as follows:
\begin{equation}
\mathbf{w}_{i,\tau} = \frac{1}{|\mathbb{C}_{i}|} (\mathbf{q}  \cdot tanh ( \mathbf{W} \cdot \mathbf{z}_i^{\tau}  + \mathbf{b} ))	  ,
\end{equation}
where $|\mathbb{C}_{i}|$ is the number of neighbor types of $\mathbf{h}_i$, $\mathbf{W}$ is the weight matrix, $\mathbf{b}$ is the bias vector, $\mathbf{q}$ is the semantic level attention vector. 
The weights would be normalized by all neighbor types:
\begin{equation}
\beta_{i,\tau} = \frac{{\rm{exp}}(\mathbf{w}_{i,\tau})}{\sum\limits_{\tau \in \mathbb{C}_{i}}  {\rm{exp}}(\mathbf{w}_{i})}  ,
\end{equation}
where $\beta_{i,\tau}$ could be interpreted as the contribution of type $\tau$ to $\mathbf{h}_i^{l}$. 
With the learned coefficients, semantic embeddings are fused to produce comprehensive embedding $\mathbf{h}_{i}^{l+1}$:
\begin{equation}
\mathbf{h}_{i}^{l+1} = \sum\limits_{\tau \in \mathbb{C}_{i}} \beta_{i,\tau} \mathbf{z}_i^{\tau} ,
\end{equation}
where $\mathbf{h}_{i}^{l+1}$ is the representation for the next HGAT layer.
Since word-type nodes possess only character-type neighboring nodes, the semantic embeddings from node-level attention could be used as representations for the next layer.

\subsection{Matcher}
To exploit the semantic clues from event labels, we convert each event label into a real-valued embedding and compute the matching score between character and event labels.
Since event labels are normally semantically related to corresponding event triggers, we initialize label embedding as the corresponding trigger-prototype character embedding.
Specifically, in data preprocessing phase, we respectively get event trigger characters of each event label in the training set of two datasets, and initialize the corresponding label embedding.
For example, label ``B-Attack'' has a set of trigger characters $\{c_1, c_2, ..., c_z\}$ in the training set, we transform these characters into char embedding, and the average value of these embeddings is used as the initial embedding of label ``B-Attack''.
So does the operation for other event labels and we formulate this process as:
\begin{equation}
\mathbf{E}_i = \frac{1}{z}\sum_{j=1}^z \bm{e}(c_{ij})  ,
\label{equ: embedding}
\end{equation}
where $\mathbf{E} \in \mathbb{R}^{k \times d}$ is the initialized label embedding matrix, which is trainable during training phase; $k$ is the number of event labels, $d$ is the dimension of label embedding,  $\mathbf{E}_i$ means the embedding of the $i_{th}$ event label, $c_{ij}$ denotes the $j_{th}$ trigger character for $i_{th}$ event label.

For a character $c_i$ and its representation $\mathbf{h}_{c_i} \in \mathbb{R}^{d}$ from the last HGAT layer, we use dot product to compute its matching score vector $\mathbf{s}_{c_i} \in  \mathbb{R}^{k}$ as:
\begin{equation}
\mathbf{s}_{c_i} = \mathbf{h}_{c_i}  \mathbf{E}^T  ,
\end{equation}
If the correct event label for $c_i$ is the  $t_{th}$ label, we denote $s_t$ as the corresponding matching score.
For other matching scores except $s_{t}$, the highest one is for the most confusing and competitive event label, and we denote it as $s_{\hat{t}}$.
In the ideal situation, we would have $s_t > s_{\hat{t}}$, which means it is easy for the model to recognize the correct event label;
However, some confusing event labels are likely to get higher matching scores than the correct label.
To discriminate confusing labels, we design a margin loss for each character $c_i$ as follows:
\begin{equation}
L_{m}(c_i) = \mathbf{max}(m + s_{\hat{t}} - s_t  , 0),
\end{equation}
where $\rm{m}$ is a positive margin.
This loss function could penalize our architecture even when $\mathbf{s}_t > \mathbf{s}_{\hat{t}} $ but the gap between $\mathbf{s}_{\hat{t}}$ and $\mathbf{s}_t$ is not large enough, thus the discriminative ability of the model is enhanced.
Sentence-level margin loss is obtained by summing margin loss of each character and we denote it as:
\begin{equation}
L_m = \sum_{i=1}^n{L_{m}(c_i)}.
\end{equation}

Recalling that we are working on a sequence labeling problem, a conditional random field (CRF) module, which is able to learn dependency relations between labels,  is employed as a sequence tagger with matching scores as inputs.
For each sentence $S = \{c_1, c_2, ..., c_n\}$, there is a corresponding label sequence $L = \{y_1, y_2, ..., y_n\}$ and a matching score matrix $\mathbf{I} \in  \mathbb{R}^{n \times k}$.
The probability of $L$ is:
\begin{equation}
P(L|S) = \frac{{\rm{exp}}(\sum\limits_{i=1}^n(\mathbf{W}_{{\rm{crf}}}^{y_i}\mathbf{I}_i +\mathbf{b}_{{\rm{crf}}}^{(y_{i-1}, y_i)}))}  {\sum\limits_{L^{'} \in \mathbb{C}} {\rm{exp}}(\sum\limits_{i=0}^n(\mathbf{W}_{{\rm{crf}}}^{y^{'}_i}\mathbf{I}_i +\mathbf{b}_{{\rm{crf}}}^{(y_{i-1}^{'}, y_i^{'})})))}   ,
\end{equation}
where $\mathbb{C}$ is the set of all arbitrary label sequences, $\mathbf{W}_{{\rm{crf}}}^{y_i}$ is the transformation matirx specific to ${y_i}$ and $\mathbf{b}_{{\rm{crf}}}^{(y_{i-1}, y_i)}$ is the transition bias specific to $(y_{i-1}, y_i)$.

We use viterbi algorithm to decode the highest scored label sequence, and get the CRF loss function for sentence $S$ as:
\begin{equation}
L_{{\rm{crf}}} = - {\rm{log}}(P(L | S)).
\end{equation}
The final optimization objective for sentence $S$ is:
\begin{equation}
L = L_{{\rm{crf}}} + \alpha L_m   ,
\end{equation}
where $\alpha$ is a hyper-parameter that controls the relative impact of margin loss and decays during training.

In test phase, we directly use the fine-tuned label embedding matrix to compute matching score between each characters between and the label embeddings, and employ viterbi algorithm to inference the highest scored label sequence.
\section{Experiments}
\begin{table*}[!ht]
	\small
	\centering
	\begin{tabular}{|l|l|c|c|c|c|c|c||c|c|c|c|c|c|}
		\hline  \multicolumn{2}{|c|}{ \multirow{3}*{ \textbf{Model} } } & \multicolumn{6}{|c||}{ACE2005} &  \multicolumn{6}{|c|}{KBP2017} \\
		\cline{3-14} \multicolumn{2}{|c|}{} & \multicolumn{3}{c|}{Trigger Identification} & \multicolumn{3}{|c||}{Trigger Classification} & \multicolumn{3}{c|}{Trigger Identification} & \multicolumn{3}{c|}{Trigger Classification} \\ 
		\cline{3-14} \multicolumn{2}{|c|}{} & \textbf{P} & \textbf{R} & \textbf{$F_1$} & \textbf{P} & \textbf{R} & \textbf{$F_1$} & \textbf{P} & \textbf{R} & \textbf{$F_1$} & \textbf{P} & \textbf{R} & \textbf{$F_1$} \\
		
		\hline \multirow{3}*{ \textbf{Feature}}& Rich-C$^{*}$    & 62.20  & 71.90 & 66.70  & 58.90 & 68.10 & 63.20  & -  & - & - & - & - & -   \\
		& KBP2017 Best   & -  & - & -  & - & - & -  & 67.76  & 45.92 & 54.74 & 62.69 & 42.48 & 50.64   \\
		
		\hline \multirow{3}*{ \textbf{Char}}& DMCNN    & 60.10  & 61.60 & 60.90  &  57.10 & 58.50  & 57.80  &  53.67 & 49.92 & 51.73  &  50.03 & 46.53 & 48.22 \\
		& C-LSTM   & 65.60  & 66.70 & 66.10 & 60.00 & 60.90 & 60.40 & -  & - & -  & - & - & -  \\
		& HBTNGMA    & 41.67  & 59.29 & 48.94  & 38.74 & 55.13 & 45.50  & 40.52  & 46.76 & 43.41  & 35.93 & 41.47 & 38.50   \\
		
		\hline \multirow{3}*{ \textbf{Word}}& DMCNN    & 66.60  & 63.60 & 65.10  & 61.60 & 58.80 & 60.20  & 60.43  & 51.64 & 55.69  & 54.81 & 46.84 & 50.51   \\
		&HNN   & 74.20  & 63.10 & 68.20  & 77.10 & 53.10 & 63.00  & -  & - & -  & - & - & -   \\
		& HBTNGMA    & 54.29  & 62.82 & 58.25  & 49.86 & 57.69 & 53.49  & 46.92  &53.57 & 50.02  & 37.54 & 42.86 & 40.03   \\
		
		\hline \multirow{3}*{ \textbf{Hybrid}} 
		& NPN & 64.8  & \textbf{73.8} & 69.0   & 60.9  & \textbf{69.3} & 64.8  & 64.32 & 53.16 & 58.21  & 57.63  & 47.63 & 52.15 \\
		& TLNN & 67.39  & 68.91 & 68.14 & 64.57 & 66.02 & 65.29 &60.5 & 56.79 & 58.59 & 59.23 & 53.11 & 56.00   \\
		& HCR (char+word) & 60.3 & 73.3 & 66.2&  58.1& 70.6 & 63.7 &- &- &- & -& -& - \\
		& HCR (char+word+lm)& 68.9  &78.8  &73.5 & 66.4 & 76.0  &70.9 &- &- &- & -& -& - \\
		\cline{2-14}
		& \textbf{HGAT(Ours)}   & 68.20 & 71.47 & 69.80 & 64.22  & 67.30 &65.73  & 61.90 & \textbf{62.84} & \textbf{62.37} & 56.48  & \textbf{57.34} &  56.90  \\
		& \textbf{L-HGAT(Ours)}  & \textbf{71.99}  & 70.83 & \textbf{71.41} & \textbf{69.38} & 68.27 & \textbf{68.82} & \textbf{63.91} & 60.06 & 61.92 & \textbf{59.21} & 55.64 & \textbf{57.37}   \\
		& \textbf{L-HGAT+BERT}  & 73.07  & 75.64 & 74.33 & 70.28 & 72.76  & \textbf{71.49}  & 69.39  & 57.75 & 63.03 & 64.37 & 53.57 &  \textbf{58.47}  \\
		
		\hline
	\end{tabular}
	\caption{\label{tab:performance} Experiment results on ACE2005 and KBPEval2017. For KBPEval2017, Trigger Identification corresponds to the Span metric and Trigger Classification corresponds to the Type metric reported in the official evaluation toolkit. ``lm'' means pretrained language model BERT.}
\end{table*}

\subsection{Datasets and Experimental Settings}
\textbf{Datasets}. 
In this paper, we conduct experiments on two popular benchmark datasets, ACE2005 and TAC KBP 2017 Event Nugget Detection Evaluation Dataset (KBP2017).
ACE2005 contains 697 articles, and we use the same data splits as~\cite{chen-ji-2009-language, feng-etal-2016-language, lin-etal-2018-nugget, ding-etal-2019-event}, where 569 articles are used for training, 64 articles for validating and the rest 64 articles for test. 
For KBP2017, we use the same setup as ~\cite{lin-etal-2018-nugget, ding-etal-2019-event}, where 506/20/167 documents are used as training/dev/test set respectively.

\textbf{Evaluation.} 
Following previous works~\cite{lin-etal-2018-nugget,ding-etal-2019-event}, we use micro-averaged Precision, Recall and $F_1$ as evaluation metrics for ACE2005, and use the official evaluation toolkit for KBP2017 to obtain these metrics.

\textbf{Hyper-Parameter Settings}. We manually tune the hyper-parameters on the dev set.
We use the same word and char embeddings as previous works~\citeauthor{lin-etal-2018-nugget}\shortcite{lin-etal-2018-nugget}.
The dimension of word embedding, char embedding, RNN representations and HGAT representations are set as 100.
Parameter optimization is performed using SGD with learning rate 0.1, L2 regularization with a parameter of 1e-5 is used to avoid overfitting. 
The max length of sentence is set to be 250 by padding shorter sentences and cutting longer ones. The number of HGAT layers is 2.
Other parameters will be listed from bottom to top in Appendix.
We run all experiments using PyTorch 1.5.1 with Python3.7 on the Nvidia Tesla 358 T4 GPU, Intel(R) Xeon(R) Silver 4110 CPU with 256GB 359 memory on Red Hat 4.8.3 OS.

\subsection{Baselines}
To comprehensively evaluate our L-HGAT model, we compare it with a series of baselines and state-of-the-art models, which could be categorized as four classes: feature-based methods, character-based NN models, word-based NN models and hybrid models.

\textbf{Feature-based methods} leverage human-designed features to conduct ED.
1) Rich-C utilizes handcraft Chinese-specific features.
2) CLUZH (KBP2017 Best) incorporates heuristic features into encoder, which once achieved the best performance in KBP2017.

\textbf{Character-based NN models} formulate Chinese ED as a character-level sequence labeling problem.
1) \textbf{DMCNN}~\cite{chen-etal-2015-event} uses dynamic multi-pooling convolution to learn sentence features for ED.
2) \textbf{C-LSTM$^{*}$}~\cite{zeng2016convolution} exploits Convolutional Bi-LSTM architecture for ED.
3) \textbf{HBTNGMA}~\cite{chen-etal-2018-collective} integrates sentence-level and document-level information through a hierarchical and bias tagging network to conduct ED.

\textbf{Word-based NN models} convert Chinese ED to a word-level sequence labeling problem.
1) The model architecture of \textbf{DMCNN} and \textbf{HBTNGMA} are the same as character-based NN models, but they are employed in the word-level.
2) \textbf{HNN}~\cite{feng-etal-2016-language} combines features extracted from CNN with Bi-LSTM to perform ED.

\textbf{Hybrid models} conduct Chinese ED from character level with incorporated word information.
1) \textbf{NPN}~\cite{lin-etal-2018-nugget} exploits character compositional structures of event triggers and utilizes gate mechanism to summarize information from character sequence and word sequence.
2) \textbf{TLNN}~\cite{ding-etal-2019-event} proposes Trigger-aware Lattice Neural Network enhanced with semantic from external linguistic knowledge base\footnote{Since we could not acquire the external sense embedding used in TLNN, we reproduce TLNN with the same Glove embedding used in this paper for fair comparison.}.
3) \textbf{HCR}~\cite{8851786} incorporates word information (position of the character inside a word and the word’s embedding) and pretrained language model, Bidirectional Encoder Representation from Transformers(BERT)~\cite{devlin-etal-2019-bert}, to improve Bi-LSTM+CRF character-wise models.
\subsection{Overall Results}
Table~\ref{tab:performance} summaries the results of L-HGAT and other baselines on both datasets, and we have analysis as follows:

(1) Our proposed L-HGAT outperforms other methods on both ACE2005 and KBP2017, which demonstrates the effectiveness of HGAT encoder and the rationality of incorporating event label embeddings.

(2) Incorporating word information into character-level models indeed helps to boost ED performance.
Character-based models are capable of alleviating word-trigger mismatch problem in theory, but Table~\ref{tab:performance} shows that word-based models outperform character-based ones, this demonstrates the effectiveness of word information.
Further, hybrid-based models surpass both character-level and word-level models by a large margin, which indicates the superiority of combing word information with character semantics.
As far as HGAT and L-HGAT, they outperform all previous methods, which manifests that our designed heterogeneous graph is efficient in exploiting the interaction between characters and lexicon words.

(3) Interaction between event labels and characters provides signals to predict event triggers more precisely.
We notice that HGAT and L-HGAT respectively gain improvements on different evaluation indicators.
HGAT shows its advantage mainly on Recall, resulting from that HGAT promotes more adequate information propagation between words and characters through our constructed heterogeneous graph, thus more potential event triggers are detected.
Meanwhile, L-HGAT further improves performance on Precision, we inference this as that the trigger-prototype-based embeddings of event labels provide semantic clues to guide classification, and margin loss enhances our model to discriminate confusing labels, leading to higher Precision.

(4) Pretrained language models help to boost performance better. 
To fully exploit the performance of our proposed model, we employ BERT, the same pretrained language model used by~\cite{8851786}, to provide contextual representations for characters. The performance of L-HGAT+BERT shows the effectiveness of pretrained language models, especially on Precision indicator. 

(5) Table~\ref{tab:performance} demonstrates that NPN achieves better recall performance on ACE2005 dataset, we inference that this is because that NPN enumerates the combinations of all characters within a window as trigger candidates, consequently more potential triggers could be predicted. 
Meanwhile, the enumeration is likely to produce invalid words, which hurt the performance on Precision. 
\section{Analysis \& Discussion}

\begin{table}[]
	\small
	\centering
	\begin{tabular}{l|c|c}
		\hline \multirow{2}*{ \textbf{Model}} &  ACE2005  & KBP2017   \\
		\cline{2-3}
		 &  \textbf{TC}  & \textbf{TC}   \\
		\hline L-HGAT  & \textbf{68.4} & \textbf{57.37} \\ 
		L-HGAT w/o $W_{\tau}$ & 66.8  & 55.9\\ 
		L-HGAT w/o c2c-edges & 65.13  & 55.26 \\
		L-HGAT w/o all-char &66.35  &  55.07 \\
		L-HGAT w/o c2w-edges & 65.33 & 56.07 \\
		L-HGAT w/o word  & 64.54  & 52.27 \\	
		\hline
	\end{tabular}
	\caption{\label{tab:hgat-compare} Experiments results on variants of L-HGAT}
\end{table}

\subsection{Comparison between Variants of L-HGAT}

HGAT is the essential encoder in our architecture. 
We intent to measure how each component of heterogeneous graph contributes to the final performance.
The experiments on variants of L-HGAT are illustrated in Table~\ref{tab:hgat-compare}, and we have analysis as follows:

1) L-HGAT w/o $W_{\tau}$ projects different types of nodes with the same convolution filter rather than node-type-specific filter, and the results decline since the heterogeneity of words and characters is not considered.

2) L-HGAT w/o c2c-edges removes connection between neighboring characters, leading to the absence of local context information.
The declination of results demonstrates that it is not enough to rely solely on word-character interaction, local context from neighboring characters also plays an important role.

3) L-HGAT w/o all-char allows the word information flowing to only the last character as Lattice-LSTM works, where information propagation between lexicon words and characters is insufficiently explored, thus the result drops approximately 2\%.

4) L-HGAT w/o c2w-edges, where sequential information of characters is not injected into words, slightly hurts the results.
This shows sequential information provides contextualized semantics for words, and better word representation can lead to better character understanding.

5) L-HGAT w/o word removes word nodes, which means only characters information is exploited.
We could see an obvious drop on results, which verifies the importance of using hybrid information from characters and lexicon words.
Besides, L-HGAT w/o word still exceeds other char-level baselines in Table~\ref{tab:performance}, which shows the effectiveness of local context and label embedding.

\begin{table}[t]
	\small
	\centering
	\begin{tabular}{l|c|c}
		\hline \multirow{2}*{ \textbf{Model}} &  ACE2005 & KBP2017 \\
		\cline{2-3} & \textbf{TC}  & \textbf{TC}  \\ 
		\hline L-HGAT & \textbf{68.82} & \textbf{57.37} \\	
		HGAT & 65.73 & 56.90 \\	
		L-HGAT w/o m-loss   & 65.56 & 55.79\\	
		L-HGAT w/o sense-emb  & 63.93  &56.20 \\
		\hline
	\end{tabular}
	\caption{\label{tab:matcher} Experiment results on variants of matcher module.}
\end{table}

\begin{CJK}{UTF8}{gbsn}
	\begin{table*}[t!]
		\small
		\centering
		\begin{tabular}{|l|c|c|c|c|c|}
			\hline \textbf{Sentence 1} &  \textbf{NPN} &  \textbf{TLNN}  &  \textbf{HGAT} &  \textbf{L-HGAT} &  \textbf{Answer}  \\
			\hline \tabincell{l}{九名逃犯被\textbf{击毙}...\\ Nine escapees were \textbf{shot} to \textbf{death}.} &  (击毙,ATTACK) & (击毙,ATTACK) &  \tabincell{l}{(击,ATTACK)\\ (毙,DIE)}   &   \tabincell{l}{(击,ATTACK)\\ (毙,DIE)}   &  \tabincell{l}{(击,ATTACK)\\ (毙,DIE)} \\ 
			\hline \hline 
			\textbf{Sentence 2} &   \multicolumn{2}{c|}{\textbf{HGAT}} &  \multicolumn{2}{c|}{\textbf{L-HGAT}} &  \textbf{Answer}  \\
			\hline \tabincell{l}{对苏珊的\textbf{指控}...\\ The \textbf{indictation} to Susan.} & \multicolumn{2}{c|}{(指控,SUE)} &  \multicolumn{2}{c|}{(指控,CHARGE-INDICT)} &  (指控,CHARGE-INDICT) \\
			\hline
		\end{tabular}
		\caption{\label{tab:case} Case study on ACE2005.}
	\end{table*}
\end{CJK}

\subsection{Comparison between Variants of Matcher}
\begin{figure}
	\centering
	\includegraphics[width=0.98\linewidth]{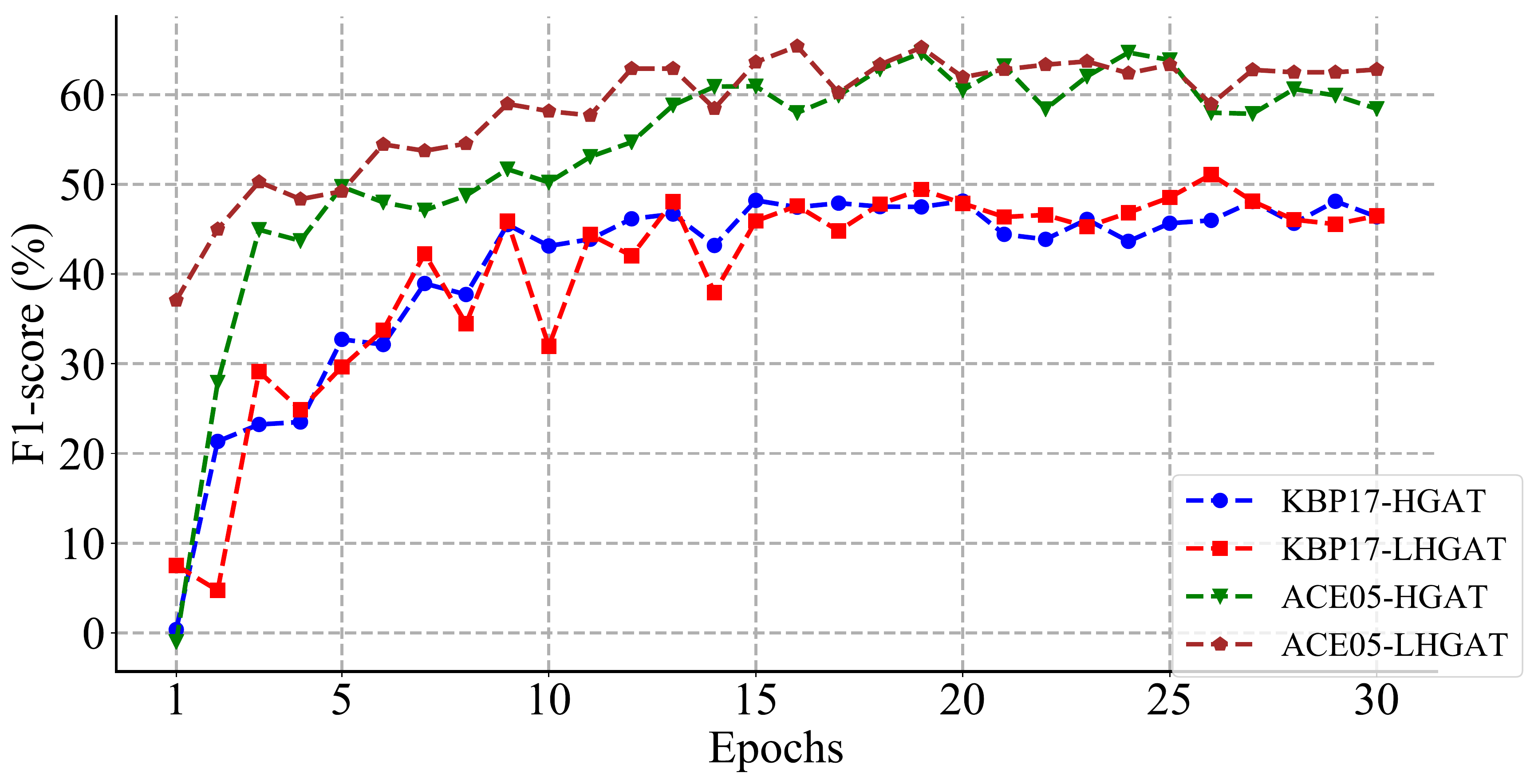}
	\caption{ $\rm{F}_1$ score of L-HGAT and HGAT on ACE2005 and KBP2017 dev set. }
	\label{fig:matcher-dev}
\end{figure}
In this section, we focus on investigating how the matcher module helps to boost performance.
Table~\ref{tab:matcher} shows that L-HGAT performs significantly better than HGAT, which demonstrates that the matcher module, including trigger-prototype based event label embedding and margin loss, provides fine-grained semantic signals to benefit ED.
Further, we have observations and analysis as follows:

(1) We probe the training process of L-HGAT and HGAT to see whether the matcher module eases the model learning process of our architecture.
As Figure~\ref{fig:matcher-dev} illustrates, L-HGAT remarkably surpasses HGAT on dev set in the early stage of training, and still maintains its advantage in the whole training process.
This phenomenon demonstrates that mining semantic information of event labels is able to provide prior knowledge to smooth the training process, guiding the detection of event triggers.

(2) We additionally notice that individually removing label embedding or margin loss performs worse than simultaneously removing them, and attribute this to two aspects:
On the one hand, trigger-prototype-based label embedding provides semantic clues to guide classification, but may be confused by event labels sharing similar sense.
In this situation, it is necessary to employ margin loss to decrease the matching score between current character and the most confusing label embedding, through which the ability to discriminate event labels is enhanced.
On the other hand, without trigger-prototype-based label embedding, the randomly initialized embedding matrix contains no semantic information, thus the margin loss may mislead the matching score between characters and the corresponding event labels.


\begin{figure}
	\centering
	\includegraphics[width=0.9\linewidth]{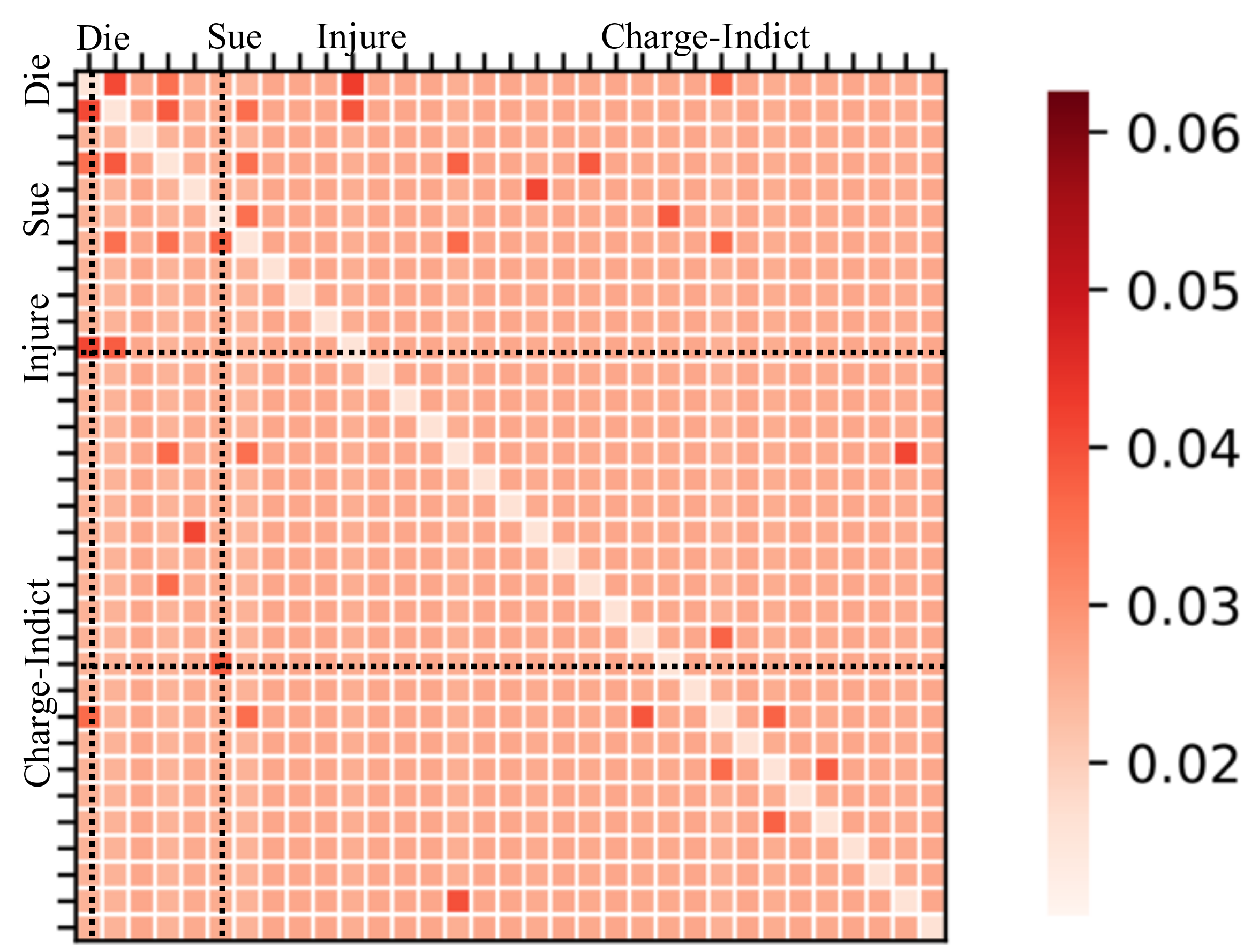}
	\caption{ Visualization of label embedding similarity(row-wise normalized) on ACE2005. We do not mark all event labels due to space issue, and the full annotation can be found in Appendix.}
	\label{fig:event-label}
\end{figure}

\subsection{Influence of Trigger Mismatch}
To further explore how word-trigger mismatch problem is alleviated with different methods, we counted the recall rate of mismatch triggers on the test set of ACE2005 and KBP2017, respectively using TLNN and HGAT.
\begin{table}[t]
	\small
	\centering
	\begin{tabular}{l|c|c}
		\hline Model &  ACE2005 & KBP2017 \\
		\hline HGAT & 92.3 & 73.63 \\
		NPN & 	84.61 & 64.55\\
		TLNN & 61.53 & 63.63 \\	
		
		\hline
	\end{tabular}
	\caption{\label{tab:mismatch} Recall rates of word-trigger-mismatch triggers on the test set of ACE2005 and KBP2017 in Trigger Identification task.}
\end{table}
Table~\ref{tab:mismatch} demonstrates that HGAT is able to handle the word-trigger mismatch better than our chosen baselines, which verifies that fuller exploitation of interaction between words and triggers help to identify the boundary of triggers much precisely. We also specifically analyze the reasons for the different performance of HGAT, TLNN and NPN with specific cases, please refer to detailed explanation in Case Study.

\subsection{Interpretability of Label Embedding}
According to the design of matcher module, we use trigger-prototype-based embedding as the initialization of event label representations, and fine-tune it during training.
To probe whether event label embeddings capture the difference and relevance between different event labels, we calculate the similarity between each pair of them.
Specifically, since each trigger must contain a ``B-EventLabel'' which marks the beginning character of event triggers, we  use the embedding of  ``B-EventLabel'' on behalf of the learned representation of the corresponding event label, and then individually calculate the cosine similarity between labels.
For clarity, we mask the diagonal score to eliminate meaningless self-similarity, and use the Softmax function to normalize the score of each line.
The visualization results are shown in  Figure~\ref{fig:event-label}.
We can observe that the similarity matrix is very sparse, since most event labels are semantically irrelevant to each other.
We further notice that some event labels carry relatively high similarity value, and these event labels share similar semantics with each other, such as (Die, Injure) and (Charge-Indict, Sue).
Therefore, we believe that label embedding is interpretable,  and capable of learning characteristic of event labels, hence providing semantic clues for ED.

\subsection{Case Study}
Table~\ref{tab:case} illustrates two examples to compare L-HGAT with other methods.
The first sentence shows that L-HGAT performs well in handling the word-trigger mismatch problem, where 
\begin{CJK}{UTF8}{gbsn}
	``击(shoot)'' and ``毙(death)'' 
\end{CJK}
are two different triggers that both are parts of 
\begin{CJK}{UTF8}{gbsn}
	the word ``击毙(shoot to death)''
\end{CJK}.
NPN gives high score to span 
\begin{CJK}{UTF8}{gbsn}
	``击毙(shoot to death)''
\end{CJK}
, which may be attributed to the similar trigger compositional structure ``verb + result'' for 
\begin{CJK}{UTF8}{gbsn}
	``打死(beat to death)'' and ``炸死(explosion to death)'' .
\end{CJK}
TLNN predicts 
\begin{CJK}{UTF8}{gbsn}
	``击毙(shoot to death)''
\end{CJK}
as trigger, we inference two reasons for this.
First, Lattice structure allows word information to flow into 
\begin{CJK}{UTF8}{gbsn}
	``毙(death)'' but ignores ``击(shoot)''
\end{CJK}
, for which the semantic information of 
\begin{CJK}{UTF8}{gbsn}
	``击(death)'' 
\end{CJK}
is not enough to recognize it as a dependent trigger.
Second, local context from neighboring characters is not fully exploited in NPN thus trigger boundary is not identified precisely.

The second sentence gives an example of how the matcher module helps to predict event labels more precisely.
As Figure~\ref{fig:event-label} demonstrates that ``SUE'' and ``INDICT'' are two semantically similar events, HGAT predicts 
\begin{CJK}{UTF8}{gbsn}
	``指控'' 
\end{CJK}
as event type ``SUE'' without the guidance of our designed matcher module.
Meanwhile, L-HGAT predicts these two labels correctly, since  trigger-prototype-based label embedding considers the interaction between triggers and event labels, and margin loss helps to discriminate confusing event labels.


\section{Conclusion}
In this paper, we propose a novel architecture, label enhanced heterogeneous graph attention networks model (L-HGAT),  for Chinese ED.
To fully exploit information between characters and words, we formulate characters and words as different types of nodes, and connect them with richly functional edges.
The heterogeneous graph attention networks is utilized to enable adequate information propagation.
Besides, we utilize the semantic clues from event labels to guide the detection of event triggers.
Experiment results show that L-HGAT consistently achieves superior performance over previous competing approaches.
In the future, we would like to adapt L-HGAT for other information extraction tasks, such as named entity recognition and aspect extraction.

\bibliography{aaai21}
\end{document}


\maketitle

\section{Hyper-parameters Settings}
We manually tune the hyper-parameters of our architecture on the dev set of two datasets respectively, and list the details of hyper-parameters in Table~\ref{tab:parameters}.
We run all experiments using PyTorch 1.5.1 with Python3.7 on the Nvidia Tesla 358 T4 GPU, Intel(R) Xeon(R) Silver 4110 CPU with 256GB 359 memory on Red Hat 4.8.3 OS.
\begin{table}[h]
	\small
	\centering
	\begin{tabular}{l|c|c}
		\toprule Parameter &  ACE2005  & KBP2017   \\
		\midrule 
		Dimension of char embeddings & 100 & 100\\ 
		Dimension of word embeddings & 100  & 100 \\
		Hidden dimension of BiLSTM units & 100  &  100 \\
		Hidden dimension of HGAT units & 100 & 100 \\
		Layers of HGAT  & 2  & 2 \\	
		$m$ (margin)  & 2  & 2 \\
		$\alpha$ (weight for margin loss)  & 0.85  & 0.85 \\	
		Layers of HGAT  & 2  & 2 \\	
		Optimizer  & SGD  & SGD \\	
		Momentum  & 0.9  & 0.9 \\	
		\bottomrule
	\end{tabular}
	\caption{\label{tab:parameters} Details of hyper-parameters settings in this paper.}
\end{table}

\section{ Visualization of label embedding similarity}
\begin{figure}[h]
	\centering
	\includegraphics[width=0.95\linewidth]{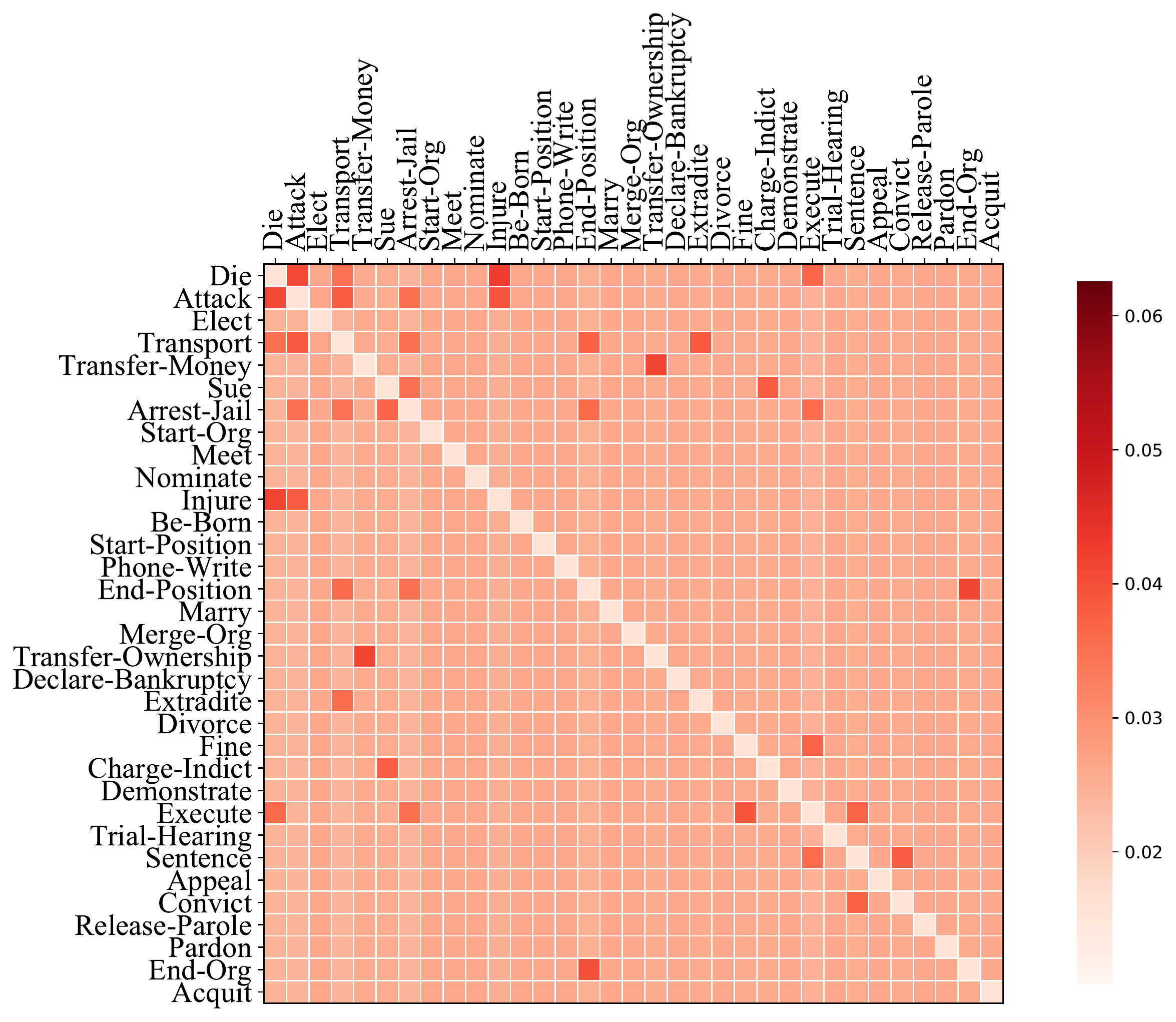}
	\caption{ Visualization of cosine similarity (row-wise normalized) between ``B-Event'' labels on ACE2005 dataset.}
	\label{fig:b-event}
\end{figure}

\begin{figure}[]
	\centering
	\includegraphics[width=0.95\linewidth]{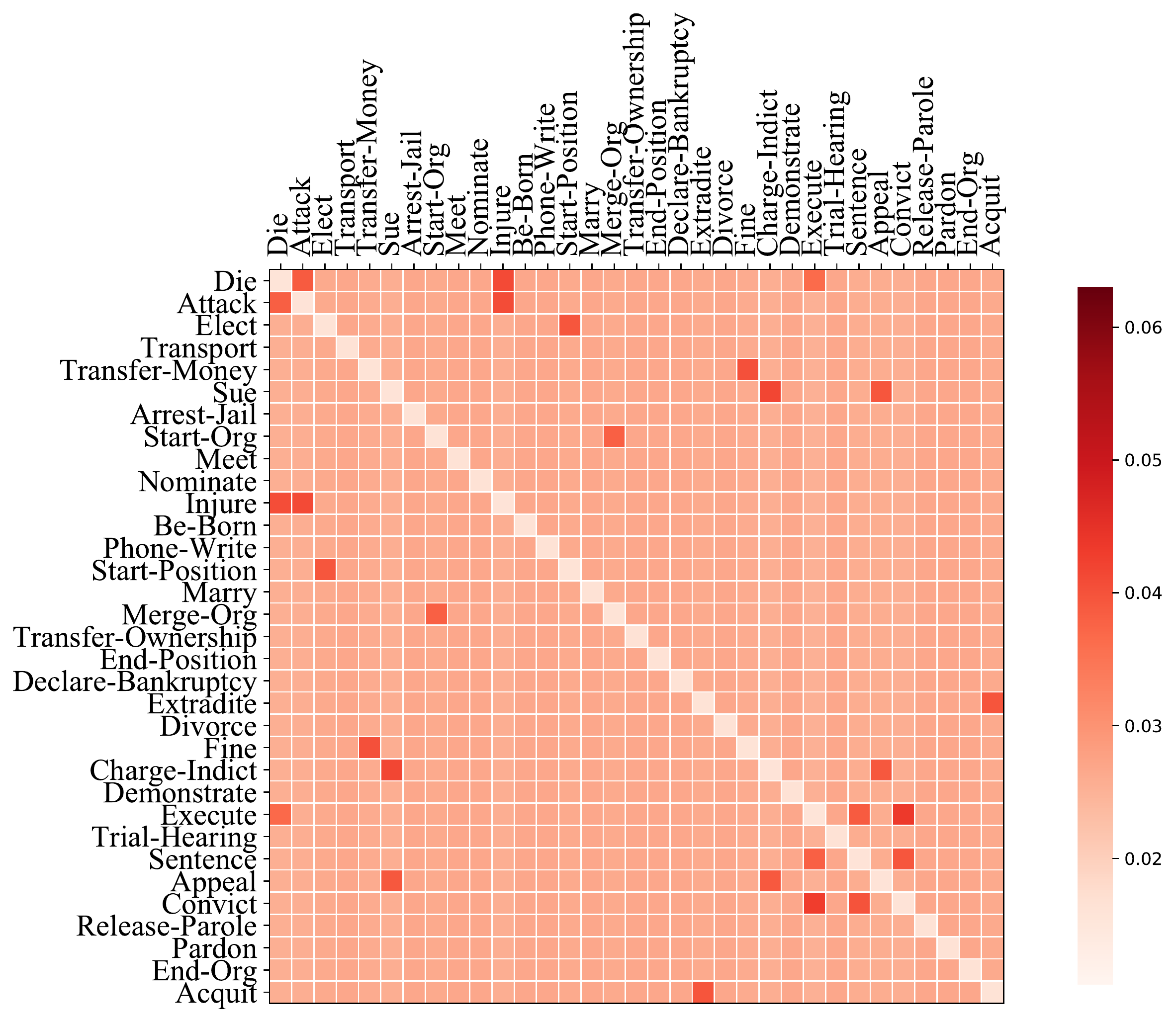}
	\caption{ Visualization of cosine similarity (row-wise normalized) between ``I-Event'' labels on ACE2005 dataset.}
	\label{fig:i-event}
\end{figure}

We show the visualization of label embedding similarity here with all event labels marked.
Figure~\ref{fig:b-event} and Figure~\ref{fig:i-event} respectively visualize the cosine similarity calculated by ``B-Event'' labels and ``I-Event'' labels.
For clarity, we mask the diagonal score to eliminate meaningless self-similarity, and use the Softmax function to normalize the value in each row.
